\documentclass{article}
\usepackage{kr}

\usepackage{times}
\usepackage{soul}
\usepackage{url}
\usepackage[hidelinks]{hyperref}
\usepackage[utf8]{inputenc}
\usepackage[small]{caption}
\usepackage{graphicx}
\usepackage{amsmath}
\usepackage{amsthm}
\usepackage{booktabs}
\usepackage{algorithm}
\usepackage{algorithmic}
\usepackage{latexsym, amsmath, amssymb, amsfonts, mathrsfs}
\usepackage{xspace}
\usepackage[T1]{fontenc}
\usepackage{float}
\usepackage{color}
\usepackage{url}
\usepackage{enumitem}
\usepackage{todonotes}
\usepackage{picinpar}
\usepackage{multirow}
\usepackage{amsthm}

\usepackage{xcolor}
\usepackage{subfig}
\usepackage{booktabs}

\usepackage{comment}
\usepackage{makecell}
\usepackage{float}
\usepackage{array}
\usepackage{placeins}

\allowdisplaybreaks

\newcommand{\KB}{\mathrm{KB}}

\newcommand{\beitemize}{\begin{list}{$\bullet$}{\topsep=1.5pt \parsep=0pt \itemsep=1pt \leftmargin=1em }} 
\newcommand{\enitemize}{\end{list}}

\newcommand{\beenumerate}{\hspace{-0.5in} \begin{enumerate}\topsep=1pt \parsep=0pt \itemsep=-3pt} \newcommand{\enenumerate}{\end{enumerate}}

\newcommand{\belist}{\begin{list}{$\bullet$}{\topsep=1.5pt \parsep=0.5pt \itemsep=1pt \leftmargin=2.25em \labelwidth=1.0em \labelsep=0.5em \partopsep=1.5pt}} 
\newcommand{\enlist}{\end{list}}

\newboolean{includeMemo}
\setboolean{includeMemo}{true} 

\newcommand{\memoside}[1]{\ifthenelse{\boolean{includeMemo}}{\todo[caption={},color=green!20!]{{\footnotesize #1}}}}
\newcommand{\memo}[1]{\ifthenelse{\boolean{includeMemo}}{\todo[inline,caption={},color=green!20!]{#1}}}
\newcommand{\memob}[1]{\ifthenelse{\boolean{includeMemo}}{\todo[inline,caption={},color=blue!20!]{#1}}}

\newcommand{\xhdr}[1]{\vspace{5pt}\noindent\textbf{#1 }}
\newcommand{\ignore}[1]{}

\newcommand{\squishlist}{
\begin{list}{{{\small{$\bullet$}}}}
{\setlength{\itemsep}{3pt}      
\setlength{\parsep}{3pt}
\setlength{\topsep}{3pt}       
\setlength{\partopsep}{3pt}
\setlength{\leftmargin}{1em} 
\setlength{\labelwidth}{1em}
\setlength{\labelsep}{0.5em} } }
\newcommand{\squishend}{  \end{list}}

\newcommand{\squishenum}{
\begin{list}{$\bullet$}{ 
    \setlength{\itemsep}{1pt}
    \setlength{\parsep}{0pt}
    \setlength{\topsep}{1.5pt}
    \setlength{\partopsep}{0pt}
    \setlength{\leftmargin}{2em}
    \setlength{\labelwidth}{1.5em}
    \setlength{\labelsep}{0.5em} } }

\newcommand{\citet}[1]{\citeauthor{#1}~\citeyear{#1}}
\newcommand{\citep}{\cite}

\renewcommand{\underline}{\uline}

\title{\textsc{TRACE-cs}: A Hybrid Logic--LLM System for Explainable Course Scheduling}

\author{%
Stylianos Loukas Vasileiou$^{1,2}$ \!\!\And\!\!\!
William Yeoh$^2$\!\\
\affiliations
$^1$New Mexico State University\\
$^2$Washington University in St. Louis\\
\emails
stelios@nmsu.edu,
wyeoh@wustl.edu
}

\begin{document}

\maketitle

\begin{abstract}
We present \textsc{TRACE-cs}, a novel hybrid system that combines symbolic reasoning with large language models (LLMs) to address contrastive queries in course scheduling problems. \textsc{TRACE-cs} leverages logic-based techniques to encode scheduling constraints and generate provably correct explanations, while utilizing an LLM to process natural language queries and refine logical explanations into user-friendly responses. This system showcases how combining symbolic KR methods with LLMs creates explainable AI agents that balance logical correctness with natural language accessibility, addressing a fundamental challenge in deployed scheduling systems.
\end{abstract}

\section{Introduction}

Scheduling systems, which allocate finite resources to multiple agents over time, are ubiquitous in real-world environments, from personnel shift assignments \cite{van2013personnel} to Mars rover activities \cite{chi2020scheduling}. Beyond generating valid and optimal schedules, it is crucial to ensure that both the schedule and the decision-making process are \textit{explainable} to human users. \textit{Explainable scheduling}, therefore, is essential for understanding scheduling decisions, rectifying issues, and providing explanations for specific decisions or schedule generation failures. Most of the work in this space have relied on symbolic, logical methods that generate valid and sound explanations.

At the other end of the spectrum, the emergence of large language models (LLMs) has marked a significant milestone in AI. While LLMs excel at generating coherent and contextually relevant text \cite{NEURIPS2020_1457c0d6}, their reliance on statistical inference leads to challenges in maintaining logical consistency and accuracy in reasoning and planning tasks \cite{mccoy2023embers,valmeekam2023planning}. This limitation is particularly apparent when explanations need to be both linguistically coherent and logically sound. In contrast, symbolic, logical methods provide a robust medium for reasoning and planning due to their ability to perform valid and sound inference. This realization offers an opportunity to combine the strengths of both LLMs and symbolic methods, creating synergistic systems that ensure decisions are not only provably correct and robust, but also communicated in a user-friendly manner.

In this paper, we present \textbf{\textsc{TRACE-cs}} (\textit{Trustworthy ReAsoning for Contrastive Explanations in Course Scheduling Problems}), a synergistic system that combines symbolic reasoning with the natural language capabilities of LLMs for generating explanations in course scheduling problems. Particularly, \textsc{TRACE-cs} generates natural language explanations for contrastive user queries (e.g., ``Why course X instead of course Y?'') by leveraging a state-of-the-art symbolic explainer \cite{vas21} together with an LLM-powered user interface for natural language interactions, thus ensuring that the explanations are provably trustworthy as well as communicated to users in a natural format. 


In short, this paper focuses on the practical implementation, deployment, and evaluation of TRACE-CS as a case study in hybrid KR systems. We demonstrate how symbolic methods provide correctness guarantees while LLMs enhance user experience through natural language processing, creating a synergistic system with real-world utility. Our experimental results quantify these benefits, showing perfect accuracy in explanations while maintaining natural language accessibility—a significant improvement over LLM-only approaches.

\begin{figure*}[!t]
    \centering
    \includegraphics[width=1.7\columnwidth]{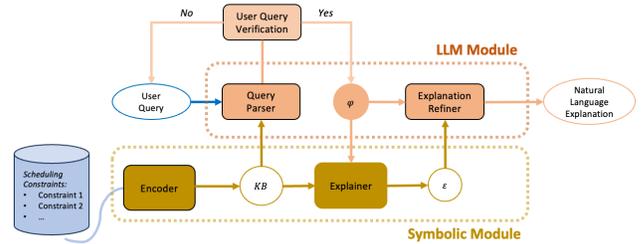}
    \caption{The \textsc{TRACE-cs} workflow.}
    \label{fig:system_architecture}
\end{figure*}

\section{Related Work}

Explainable scheduling research has predominantly relied on logical symbolic methods \cite{Cyras+19,agrawal2020using,bertolucci2021explaining,pozanco2022explaining,powell2022abstract,vasLMRP,vasileioua2023lasp,zehtabi2024contrastive}. While grounded in sound inference procedures, these approaches often produce explanations that are difficult to communicate to users due to their logic-based nature. Attempts to mitigate this limitation have used templates mapping logical explanations to pre-specified natural language sentences \cite{pozanco2022explaining,vasileioua2023lasp} or visualization interfaces \cite{vcyras2021schedule,kumar2022vizxp,powell2022abstract}. 

Concurrently, LLMs have revolutionized natural language processing and found applications across diverse domains, including planning \cite{kambhampati2024llms}, code generation \cite{roziere2023code}, and medical applications \cite{zhou2023survey}. However, the integration of LLMs with symbolic explainable scheduling systems remains largely unexplored. Our work, \textsc{TRACE-cs}, represents the first attempt to address this gap by presenting a novel hybrid system that synergistically combines a symbolic explainable scheduling module with an LLM module.

\section{\textsc{TRACE-cs} System Architecture Overview}

The \textsc{trace-cs} system architecture, illustrated in Figure \ref{fig:system_architecture}, consists of two primary components: a \textit{Symbolic Module} handling constraint encoding and explanation generation, and an \textit{LLM Module} managing natural language interaction. The workflow is as follows: (1) The user submits a contrastive query in natural language; (2) The Query Parser extracts the information from the query and converts it into a logical representation $\varphi$ consistent with the knowledge base $\KB$ created by the Encoder; (3) The user verifies if the extracted query information corresponds to the original query, and proceeds to the next step if it is; (4) The Explainer generates a symbolic explanation $\epsilon$ for $\varphi$ with respect to $\KB$; (5) The Explanation Refiner converts $\epsilon$ into natural language and outputs it to the user.

\subsection{Symbolic Module} 

The Symbolic Module forms the core of \textsc{TRACE-cs} and consists of two subcomponents, the \textit{Encoder}, which encodes the scheduling constraints into a logical knowledge base, and the \textit{Explainer}, which generates minimal explanations for user queries with respect to the knowledge base.

\xhdr{Encoder.} The Encoder transforms course scheduling constraints into a Boolean satisfiability (SAT) problem.\footnote{A plethora of scheduling problems has been modeled using SAT-based approaches \cite{crawford1994experimental,pinto1997logic,kundu2008sat,ansotegui2011satisfiability,haspeslagh2013efficient,bofill2015maxsat,demirovic2019modeling}. In such problems, a schedule is found if and only if the encoded $\KB$ has a satisfying model.}  The system models scheduling decisions using Boolean variables and logical clauses derived from degree requirements and university policies. Specifically, for each course $c$ and semester $s$, the system creates a course variable $\text{var}(c,s)$ that indicates whether course $c$ is scheduled in semester $s$. The constraints span several categories, such as degree requirements (e.g., core courses, elective distributions, total credit requirements), temporal constraints (e,g., prerequisites, semester credit limits), and general scheduling constraints (e.g., each selected course is assigned to exactly one semester). For example, the prerequisite constraint `YNP H57 must be completed before XOX R89'' is encoded as the following logical clause: $\neg \text{var}(\text{XOX\_R89}, s) \vee \bigvee_{t=0}^{s-1} \text{var}(\text{YNP\_H57}, t)$, where $s$ is a semester that XOX R89 could be scheduled, and $t < s$. This ensures that if XOX R89 is scheduled in semester $s$, then YNP H57 must be scheduled in some previous semester $t < s$.

\xhdr{Explainer.}The Explainer generates a minimal explanation for the user contrastive query $\varphi$ (processed by the LLM module) using the logic-based explanation generation algorithm from \cite{vas21,vasileioua2023lasp}. In essence, the algorithm takes as input the $\KB$ and the query $\varphi$, where $\KB \models \varphi$, and outputs a set of logical clauses $\epsilon \subseteq \KB$ such that $\epsilon \models \varphi$, and $\nexists \epsilon' \subset \epsilon$ such that $\epsilon' \models \varphi$.\footnote{Practically, the algorithm leverages the fact that if $\KB \models \varphi$, then $\KB \wedge \neg \varphi \models \bot$, and uses SAT-based solvers optimized to find minimal unsatisfiable sets (MUSes) and minimal correction sets (MCSes) \cite{marques2012computing,marques-silva-ijcai13}.} In other words, it outputs a $\subseteq$-minimal explanation $\epsilon$.

  \begin{figure*}[!t]
    \centering
    \includegraphics[width=\textwidth]{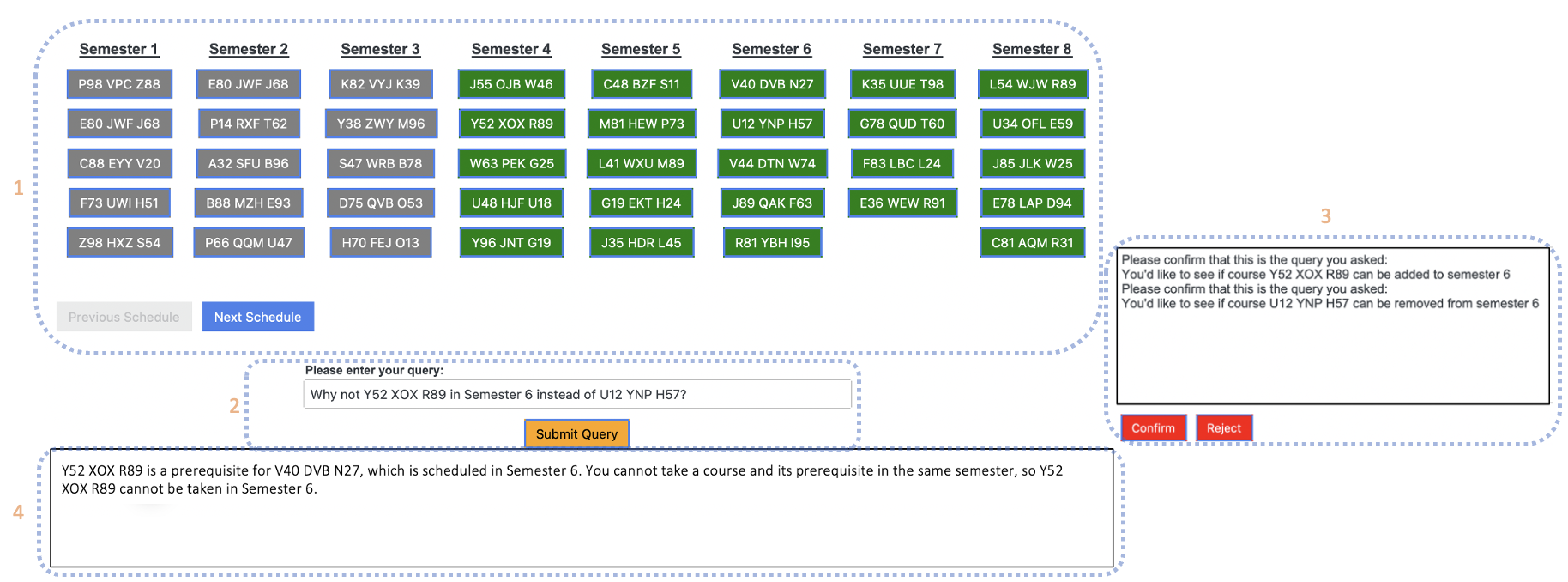}
    \vspace{-1em}
    \caption{The \textsc{TRACE-cs} interface showing the explanation workflow: (1) course schedule display across 8 semesters; (2) user query input; (3) query verification step with user confirmation; and (4) generated explanation output. }
    \label{fig:interface}
\end{figure*}

\subsection{LLM Module}
The LLM Module serves as the interface between the user and the Symbolic Module, handling natural language processing tasks through two subcomponents: the \textit{Query Parser}, which interprets contrastive queries and converts them into logical representations, and the \textit{Explanation Refiner}, which translates logical explanations into user-friendly natural language responses.

\xhdr{Query Parser.} The Query Parser converts natural language contrastive queries into logical representations $\varphi$ compatible with the knowledge base $\KB$. The parser uses an LLM with in-context learning to extract three key components from user queries: (1) course names, (2) target semesters, and (3) conditions (positive or negative). For example, the query ``Why not XOX R89 instead of YNP H57?'' is parsed to extract two courses: XOX R89 with a positive condition (should be scheduled) and YNP H57 with a negative condition (should not be scheduled), both targeting the semester where YNP H57 is currently scheduled.\footnote{The parser includes examples in the prompt to handle various query formats and performs fuzzy matching for partially specified course names.} The extracted information is then converted into clauses that can be evaluated against the knowledge base. Finally, as recent work highlights potential limitations of LLMs in formal interpretation tasks \cite{karia2024can}, \textsc{TRACE-cs} includes a user verification step before proceeding to explanation generation to ensure that the converted queries are correct.

\xhdr{Explanation Refiner.} The Explanation Refiner takes the symbolic explanation $\epsilon$ from the Explainer and translates it into coherent natural language responses. For each clause in $\epsilon$, the system retrieves the corresponding English label—a short sentence describing what the constraint means in natural language. The refiner then uses an LLM with in-context learning to process these labels along with contextual information about the current schedule and course descriptions, generating a coherent explanation while maintaining logical accuracy. For instance, multiple related constraints about prerequisite violations are grouped and presented as a single coherent explanation rather than as separate constraint statements.

\section{Proof-of-Concept}

We implemented \textsc{TRACE-cs} as a proof-of-concept for undergraduate computer science course scheduling at our university.\footnote{Code repository: \url{https://github.com/YODA-Lab/TRACE-CS}.} The system handles scheduling decisions across eight academic semesters, incorporating real course data and degree requirements from the university's official sources.

\xhdr{Domain and Data Collection.}The application domain involves scheduling courses for a Bachelor of Science in Computer Science degree, which requires 120 credit hours distributed across core courses, electives, and general education requirements. We collected course data by scraping our university's official course catalog and degree requirements, extracting course codes, credit hours, prerequisites, and course descriptions. The dataset includes over 200 courses spanning core CS courses, CS electives, science electives, and social science/humanities requirements. All courses are anonymized for the blind review process. Prerequisites form complex dependency chains--for instance, WJW R89 (Analysis of Algorithms) requires XOX R89 (Data Structures), which in turn requires VPC Z88 (Introduction to Computer Science).

\xhdr{Implementation.}The system is implemented in Python, with the Symbolic Module using the PySAT library \cite{imms-sat18} for SAT encoding and solving, and the LLM Module using GPT-4.1 \cite{openAI2023} for natural language processing. The encoder generates constraints for degree requirements (e.g., ``at least 45 CS elective credits''), temporal dependencies (e.g., prerequisite chains), and scheduling logistics (e.g., 9-15 credits per semester). The system produces multiple valid schedules using solution blocking techniques, allowing users to explore different scheduling options.

\xhdr{User Interface.}Figure~\ref{fig:interface} shows the interface, which displays the generated schedule as a semester-by-semester course layout. Users can submit contrastive queries through a text input field, such as ``Why not XOX R89 in semester 6 instead of YNP H57?''. The interface includes a verification step where users confirm that the system correctly parsed their query before proceeding to explanation generation. This verification step addresses potential limitations in LLM query interpretation and ensures user intent is accurately captured.

\xhdr{Query Types and System Response.}The system handles various contrastive query patterns, including single-course questions (e.g., ``Why XOX R89?''), temporal queries (``Why not XOX R89 in semester 5?''), and comparative queries (``Why LAP D94 instead of UUE T98 in Semester 6?''). For each query, the system identifies a minimal set of constraints preventing the alternative and presents explanations such as ``WJW R89 cannot be scheduled because its prerequisite XOX R89 has not been completed" or ``The total credits for CS electives must sum to 45 credits.'' The interface maintains conversation history, allowing users to ask follow-up questions about the same schedule.

\begin{table}[t]
\centering
\resizebox{1.\columnwidth}{!}{ 
\begin{tabular}{c|cc|cc|cc}
\hline
Complexity & \multicolumn{2}{c|}{Accuracy (\%)} & \multicolumn{2}{c|}{Avg. Words} & \multicolumn{2}{c}{Avg. Runtime (sec)} \\
Level  & \textsc{TRACE-cs} & GPT-4.1 & \textsc{TRACE-cs} & GPT-4.1 &\textsc{TRACE-cs} & GPT-4.1 \\
\hline
1 & 100.0 & 62.0 & 64.7 & 129.8  & 11.0 & 3.9\\
2 & 100.0 & 56.0 & 63.0 & 152.1  & 12.2 & 4.5\\
4 & 100.0 & 52.0 & 102.7 & 165.3 & 15.9 & 4.2  \\
6 & 100.0 & 46.5 & 122.4 & 190.5 & 20.0 & 4.7  \\
\hline
\hline
Overall& \textbf{100.0} & \textbf{54.1} & \textbf{81.8} & \textbf{159.4} & \textbf{14.7} & \textbf{4.3} \\
\hline
\end{tabular}
}
\caption{Comparative evaluation results between \textsc{TRACE-cs} and GPT-4.1 approach across 550 queries of various complexity levels.}
\label{tab:results}
\end{table}

\subsection{Computational Evaluation}

We conducted a comparative evaluation of \textsc{TRACE-cs} against a pure LLM-only approach using GPT-4.1.\footnote{We chose GPT-4.1 because it was one of the best performing and most affordable model at the time of writing this paper.} The evaluation used 550 contrastive queries across several different course schedules, and was run on a machine with an M1 Max processor and 32GB of RAM.

\xhdr{Experimental Setup.}We generated queries of varying complexity levels, where complexity indicates the number of courses mentioned in the query (e.g., ``Why not YNP H57?'' has complexity 1, while ``Why VPC Z88 in semester 1 and JWF J68 in semester 2?'' has complexity 2). For the LLM-only baseline, we provided GPT-4.1 with the course schedule, course descriptions, and all scheduling constraints, asking it to generate explanations directly without the logical reasoning component. We also provided it with a few example queries and their correct corresponding explanations.

\xhdr{Evaluation Metrics.}We measured three key aspects: (1)~\textit{explanation correctness} with respect to the scheduling constraints, evaluated manually by the authors, (2) \textit{verbosity} measured by word count in generated explanations, and (3) \textit{response time} for explanation generation.

\xhdr{Results.}Table~\ref{tab:results} shows the results. \textsc{TRACE-cs} achieved 100\% correctness across all complexity levels, while GPT-4.1 achieved only 54.1\% correctness overall, with performance ranging from 62.0\% for queries of complexity 1 to 46.5\% for queries of complexity 6. The low accuracy indicates a limitation of pure LLM approaches for logical reasoning tasks. In terms of verbosity, GPT-4.1 generated significantly longer explanations, averaging 159.4 words compared to \textsc{TRACE-cs}'s 81.8 words. This verbosity increased substantially with query complexity, reaching 190.5 words for complexity 6 queries compared to \textsc{TRACE-cs}'s 122.4 words. It is worth noting that GPT-4.1 exhibited a tendency to generate non-minimal explanations that included most or all applicable constraints rather than identifying the specific minimal set causing the conflict, despite being prompted to only generate the most relevant and minimal reasons. While these comprehensive explanations may be technically correct in some cases, they might overwhelm users as they include unnecessary information.

For response time, as expected, GPT-4.1 was faster, averaging 4.3 seconds compared to \textsc{TRACE-cs}'s 14.7 seconds. However, \textsc{TRACE-cs}'s additional computational cost (due to calling SAT solvers) leads to perfect accuracy of the explanation. Overall, the results reveal a critical trade-off between speed and reliability in explanation generation. GPT-4.1's poor performance possibly stems from its statistical inference approach, which struggles with the precise logical reasoning required for constraint satisfaction problems. In contrast, \textsc{TRACE-cs} leverages symbolic reasoning to guarantee logical correctness while using the LLM component solely for natural language processing tasks where it excels.

\section{Conclusions}

We presented \textsc{TRACE-cs}, a hybrid system that combines logical reasoning with LLMs for explainable course scheduling. Our evaluation demonstrates that \textsc{TRACE-cs} achieves perfect logical correctness while generating concise, minimal explanations--substantially outperforming the pure LLM approach that achieved only 54.1\% accuracy. As LLM capabilities continue to evolve, the modular design of \textsc{TRACE-cs} might provide a framework for incorporating improved reasoning models while maintaining the guarantee of logical correctness through symbolic verification.

It is important to note that our evaluation focused on a single LLM (i.e., GPT-4.1). Recent advances in reasoning capabilities of LLMs suggest that newer or more specialized models might achieve better performance on logical reasoning tasks. Models specifically trained on formal reasoning or those with enhanced chain-of-thought capabilities could potentially narrow the gap with symbolic approaches. Evaluating the system with newer models and/or domain-specific reasoning models would provide insights into the evolving landscape of neural reasoning capabilities. Additionally, our evaluation used a straightforward prompting strategy; more sophisticated prompting techniques, such as structured reasoning prompts or multi-step verification processes, might improve LLM performance as well. 

Finally, while demonstrated on course scheduling, the hybrid architecture of \textsc{TRACE-cs} can extend naturally to other constraint-based domains, such as planning and resource allocation. Moreover, conducting user studies to assess explanation quality from an end-user perspective would provide valuable insights into the practical utility of minimal versus comprehensive explanations in real-world deployment scenarios.

\section*{Acknowledgements}

Stylianos Loukas Vasileiou and William Yeoh are partially supported by the National Science Foundation (NSF) under award 2232055. The views and conclusions contained in this document are those of the authors and should not be interpreted as representing the official policies, either expressed or implied, of the sponsoring organizations, agencies, or governments.

\bibliographystyle{kr}
\bibliography{refs}

\end{document}